\pdfoutput=1

\documentclass[11pt]{article}

\usepackage[preprint]{acl}

\usepackage{times}
\usepackage{latexsym}
\usepackage{listings}
\usepackage{amsmath}
\usepackage{enumitem}
\usepackage{booktabs}
\usepackage{tabularx}

\usepackage[T1]{fontenc}

\usepackage[utf8]{inputenc}

\usepackage{microtype}

\usepackage{inconsolata}

\usepackage{graphicx}
\usepackage{booktabs}
\usepackage{xcolor}
\usepackage[normalem]{ulem}
\useunder{\uline}{\ul}{}
\usepackage{tabularx}
\usepackage{amsmath}
\newcolumntype{Y}{>{\centering\arraybackslash}X}

\title{How Instruction-Tuning Imparts Length Control: A Cross-Lingual Mechanistic Analysis}

\author{
\textbf{Elisabetta Rocchetti\textsuperscript{1}},
\textbf{Alfio Ferrara\textsuperscript{1}}
\\
\textsuperscript{1}Università degli Studi di Milano, Department of Computer Science, Via Celoria, 18 - 20133 Milan, Italy\\
\small{
\textbf{Correspondence:} name.surname@unimi.it
}
}

\begin{document}
\maketitle

\begin{abstract}
 Adhering to explicit length constraints, such as generating text with a precise word count, remains a significant challenge for Large Language Models (LLMs). This study aims at investigating the differences between foundation models and their instruction-tuned counterparts, on length-controlled text generation in English and Italian. We analyze both performance and internal component contributions using Cumulative Weighted Attribution, a metric derived from Direct Logit Attribution. Our findings reveal that instruction-tuning substantially improves length control, primarily by specializing components in deeper model layers. Specifically, attention heads in later layers of IT models show increasingly positive contributions, particularly in English. In Italian, while attention contributions are more attenuated, final-layer MLPs exhibit a stronger positive role, suggesting a compensatory mechanism. These results indicate that instruction-tuning reconfigures later layers for task adherence, with component-level strategies potentially adapting to linguistic context.
\end{abstract}

\section{Introduction}
Large Language Models (LLMs) are increasingly integrated into diverse applications, from rephrasing sentences for formality to summarizing texts within strict character limits. This widespread adoption underscores the growing necessity for LLMs to accurately interpret and adhere to user-specified instructions embedded within prompts. Instruction-tuning has emerged as a key technique to enhance this capability, aligning LLM outputs more closely with user expectations, often in a conversational or chat-like format.

The development of instruction-tuned models has spurred research into the underlying mechanisms that enable such adaptability, as well as systematic evaluations of their performance on various control tasks~\cite{liang2024controllabletextgenerationlarge}. These benchmarks have revealed that LLM proficiency varies significantly depending on the nature of the instruction. For example, while adjusting to a specific tone is generally well-handled by current LLMs, generating text that strictly adheres to explicit length constraints—such as a specified number of words, characters, or syllables—remains a notably challenging task~\cite{zhou2023instructionfollowingevaluationlargelanguage, lu-etal-2023-bounding, sun-etal-2023-evaluating, chen2024benchmarking, chen2024evaluatingunderstandingimprovingconstrained, yao2024collie}.

This study focuses specifically on the word count constraint, conducting an in-depth analysis of the Llama 3.1 8B model~\cite{grattafiori2024llama3herdmodels}. We investigate both its performance and the contributions of its internal components to this task. By employing Mechanistic Interpretability techniques, particularly Direct Logit Attribution and our derived Cumulative Weighted Attribution (CWA) metric, we aim to pinpoint which model components (e.g., attention heads, MLPs) are pivotal for successfully generating text of a requested length. Our analysis features a direct comparison between a foundation model and its instruction-tuned counterpart, shedding light on how fine-tuning impacts these internal mechanisms. Furthermore, we enrich this investigation with a cross-lingual comparison, examining model behavior and component contributions in both English and Italian to explore the language-specificity of these instruction-following capabilities.

The remainder of this paper is structured as follows. We first review related work on controllable generation and interpretability (Section~\ref{sec:related_work}) and introduce preliminaries on auto-regressive models and DLA (Section~\ref{sec:preliminaries}). We then describe our methodology, including the Cumulative Weighted Attribution (CWA) metric, and our experimental setup (Sections~\ref{sec:methodology} and \ref{sec:exp-setup}). In Section~\ref{sec:results}, we present our findings on the length control capabilities of Llama 3.1 8B and use CWA to analyze internal component contributions. Finally, Section~\ref{sec:conclusion} summarizes our findings, discusses limitations, and suggests directions for future research.

\section{Related Work}
\label{sec:related_work}
Equipping LLMs with the ability to adapt to users' needs and constraints is essential for improving their utility and reliability. This section reviews the landscape of Controllable Text Generation, current limitations in output length control, relevant interpretability techniques, and the impact of instruction tuning, all of which provide context for our investigation into multilingual word count control.
\paragraph{Controllable Text Generation}
LLMs are often required to produce outputs which adheres to specific instructions or constraints. This is referred to as Controllable Text Generation (CGT), and encompasses a wide variety of control specifications. 
\citet{liang2024controllabletextgenerationlarge} provides a comprehensive taxonomy distinguishing between explicit and implicit control. 

\textit{Explicit} control involves defining constraints in the input prompt expressed in natural language, while \textit{implicit} control refers to producing content which meets quality standards and ethical guidelines. 
Concerning explicit control tasks, the main distinction is made between \textit{hard} control and \textit{soft} control: the former focuses on contraints related to the output's structure or vocabulary (e.g. JSON format, text length, keyword inclusion/exclusion); the latter includes instructions on abstract attributes such as sentiment or style (e.g. positive sentiment, Shakespearean style). 
This work focuses on the analysis of LLMs' performance in \textit{word count control}, which is an explicit hard constraint often required in practical applications yet demonstrably challenging for current models.
\paragraph{Length Control Capabilities}
Both open-source and commercial LLMs' performances have been tested on length-controlled text generation task. It has been found that most LLMs show poor capabilities in generating texts with the requested length~\cite{zhou2023instructionfollowingevaluationlargelanguage, lu-etal-2023-bounding, sun-etal-2023-evaluating, chen2024benchmarking, chen2024evaluatingunderstandingimprovingconstrained, yao2024collie}, with commercial models (e.g. GPT-3.5~\cite{gpt35}, GPT-4~\cite{openai2024gpt4technicalreport}) performing better than open-source models (e.g. LLaMA2~\cite{touvron2023llama2openfoundation}, Vicuna~\cite{vicuna2023}, Mistral~\cite{jiang2023mistral7b}). 
Notably, performance tends to degrade as the target word count increases~\cite{lu-etal-2023-bounding, sun-etal-2023-evaluating, yao2024collie}.

\citet{yao2024collie} investigated a wider range of prompt formulations for length-controlled text generation, including the classical template \textit{Generate a sentence with N words.} as well as more flexible variants, such as \textit{Generate a sentence with at least N words.} Their findings indicate that models struggle more with prompts involving stricter counting or positional constraints. While GPT-4 showed some improvement in handling these cases, its performance remained far from perfect. 
Even with few-shot in-context learning, performance remains low~\cite{sun-etal-2023-evaluating}.

One reason for the poor performance on length-controlled text generation is that LLMs lack an explicit counting mechanism. Furthermore, since language models operate on tokens rather than words, it becomes even more challenging for them to track the number of words generated from their own perspective~\cite{chen2024benchmarking}.

To the authors' knowledge, there is currently no work in the literature considering the comparison across multiple languages on length-controlled text generation.
\paragraph{Approaches to Understanding LLM Control}
Investigating how LLMs handle control tasks, particularly adherence to length constraints, centers on understanding their internal mechanisms and how these are shaped by processes like instruction tuning. Such insights are foundational for any subsequent efforts to improve performance.

One avenue for gaining this internal understanding is Mechanistic Interpretability (MI)~\cite{rauker2023transparentaisurveyinterpreting, miolah}, which seeks to reverse-engineer model computations into interpretable components or circuits. Identifying model components that contribute most to satisfying length constraints, for instance, offers a potential path towards improving this capability. 
A commonly used technique in MI is direct logit attribution (DLA), which quantifies the contribution of individual components to the output logits, and ultimately, to task success. DLA has been used to identify components playing a relevant role in several linguistic tasks, e.g. subject-verb agreement~\cite{yin-neubig-2022-interpreting, ferrando2024onthesimilarity}. 
In particular,~\citet{ferrando2024onthesimilarity} employed DLA to compare component contributions in Gemma 2B~\cite{gemmateam2024gemmaopenmodelsbased} across English and Spanish.
Building on this, our work will employ DLA to pinpoint components crucial for word count control and investigate whether these components differ across languages.

A comprehensive understanding of LLM control involves examining how training paradigms like instruction tuning (IT) alter model behavior and potentially shift the roles of internal components. Comparing instruction-tuned (IT) models with their foundational counterparts can reveal how chat models adapt during the fine-tuning process—specifically, how they align their outputs to better respond to user prompts, which often include explicit instructions like those for length. 
Recent work has examined the outcomes of instruction tuning and its impact on specific model components~\cite{kung2023models, gao-etal-2023-roles, wu-etal-2024-language}. 
Several studies suggest that instruction-tuned models primarily learn superficial patterns, such as mimicking output formats or relying on prompt cues to infer expected responses~\cite{kung2023models, zhou2023instructionfollowingevaluationlargelanguage}. 
Conversely, IT has also been shown to enable models to recognize instruction words in user prompts and to consistently condition their generation on these cues, without necessarily altering the underlying linguistic capabilities of the pre-trained model. Notably, instruction tuning can strengthen attention connections related to instruction verbs~\cite{wu-etal-2024-language}.
Therefore, comparing foundation and instruction-tuned models is crucial for understanding how the ability to follow length constraints (or the lack thereof) emerges or changes during the fine-tuning process, and whether these changes are influenced by language.
\section{Preliminaries}
\label{sec:preliminaries}
In this section, we introduce the fundamental concepts and notation used throughout the remainder of the paper. We begin by defining the language modeling objective and describing how Transformer-based autoregressive LLMs compute it. We then present the Direct Logit Attribution method, which we use to identify task-relevant components within the model.
\paragraph{Auto-regressive LLMs}
We can express the auto-regressive language modeling process performed by an LLM as the probability of a token sequence $T = (t_1, t_2, \dots, t_K)$:
\begin{equation}
\label{eq:lm}
  P(T) = P(t_1, t_2, \dots, t_K) = \prod_{i = 1}^{K}p(t_i|t_{<i})
\end{equation}
where $t_i$ represents the token at position $i$, $t_{<i}$ denotes the sequence of preceding tokens $(t_1, \dots, t_{i-1})$, and $K$ is the total number of tokens in the generated sequence $T$.

In a Transformer-based LLM, $p(t_i|t_{<i})$ is derived from the final hidden state as computed during the $i$-th generation step. The model architecture consists of an embedding layer, $L$ stacked Transformer blocks, and a final output layer.
Let $\mathbf{h}^{(0)}$ be the input token embeddings (plus positional embeddings). The $l$-th Transformer block, denoted $b_l$, processes an input $\mathbf{h}^{(l-1)}$ (the output from the preceding block) to produce $\mathbf{h}^{(l)}$. Each block $b_l$ typically comprises a multi-head self-attention module ($A_l$, taking as input $\mathbf{h}^{(l-1)}$), a feed-forward network ($F_l$, taking as input $\mathbf{h}^{(l-1)}_{\text{\textit{post-Attn}}}$, which is the output of $A_l$), and layer normalization ($LN$, which takes as input $F_l$'s output). Each block computes $\mathbf{h}^{(l)}$ as 
\begin{align*}
  b_l(\mathbf{h}^{(l-1)}) = LN(F_l(\mathbf{h}^{(l-1)}_{\text{\textit{post-Attn}}}) + \mathbf{h}^{(l-1)}_{\text{\textit{post-Attn}}})\\
  \mathbf{h}^{(l-1)}_{\text{\textit{post-Attn}}} = LN(A_l(\mathbf{h}^{(l-1)}) + \mathbf{h}^{(l-1)})
\end{align*}
The final output layer then takes $\mathbf{h}^{(L)}$ and projects it back to the vocabulary space using the unembedding weight matrix $W_U$:
\begin{math}
  \mathbf{h}^{(L)}W_U
\end{math}.
\paragraph{Direct Logit Attribution}
As noted in~\citet{elhage2021mathematical}, while individual components like attention ($A_l$) and feed-forward networks ($F_l$) contain non-linearities, their outputs are added to the residual stream. The final hidden state $\mathbf{h}^{(l-1)}$ is thus a sum of the initial embedding and the outputs of all intermediate attention and feed-forward network modules. This additive structure in the residual stream allows for computing the direct effect of a component to the final logits, which we call Direct Logit Attribution (DLA).
DLA for attention modules can be easily computed as
\begin{equation}
\label{eq:dla-attn}
  \text{DLA}_{A_l} = A_l(\mathbf{h}^{(l-1)})W_U
\end{equation}
whereas the DLA for feed-forward networks can computed as
\begin{equation}
\label{eq:dla-mlp}
  \text{DLA}_{F_l} = F_l(\mathbf{h}^{(l-1)}_{\text{\textit{post-Attn}}})W_U.
\end{equation}
\section{Methodology}
\label{sec:methodology}
\begin{figure}
  \centering
  \includegraphics[width=\linewidth]{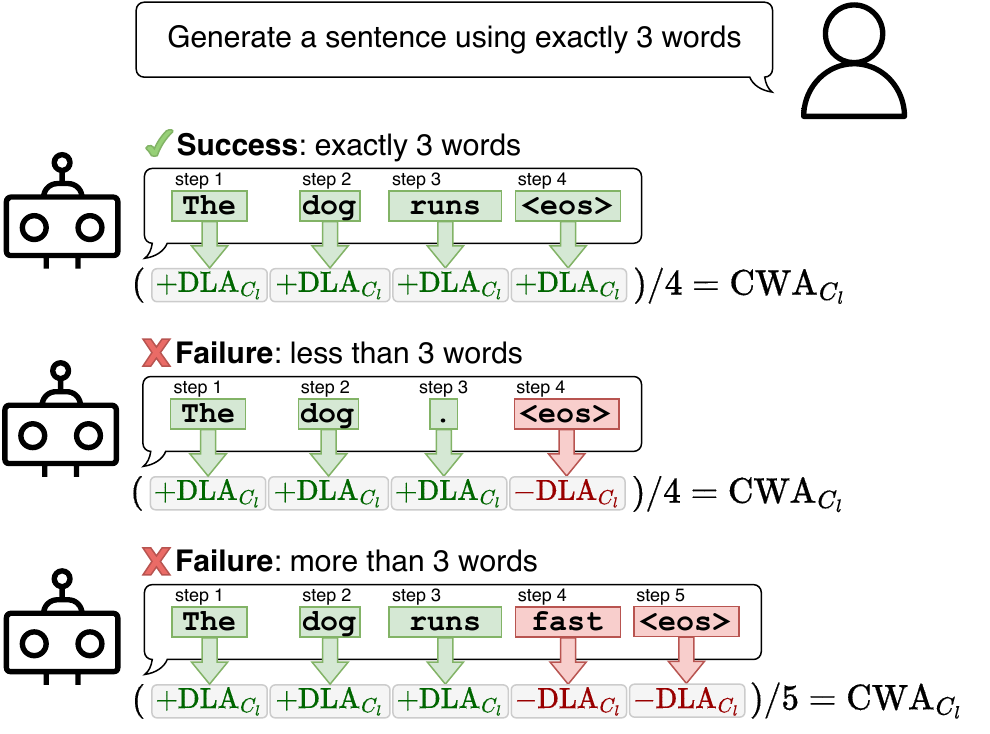}
  \caption{Success and failure cases with corresponding $\text{CWA}_{C_l}$ calculations. DLA scores at each generation step are aggregated with a positive or negative sign, depending on whether the produced token adheres to the specified word count constraint.}
  \label{fig:example}
\end{figure}

Our objective is to identify model components that are critically involved when an LLM attempts to satisfy an explicit word count constraint of $N$ words. We employ DLA to analyze an instruction-tuned LLM's behavior when prompted with instructions like: ``\textit{Generate a sentence using exactly $N$ words.}'' (e.g., for $N=3$, ``\textit{Generate a sentence using exactly 3 words.}''). The model's output sequence $T$ can result in three primary scenarios relative to the target word count: success, failure due to early termination, and failure due to late termination (see Figure~\ref{fig:example} for a visual depiction of these three cases).

\textbf{Success (Correct Length)}: The model generates exactly $N$ words before the end-of-sentence (EOS) token.
\textit{Example ($N=3$):} \texttt{The dog runs <eos>}.
In this scenario, we consider all token generation steps contributing to the $N$ words as ``correctly contributing'' to satisfying the constraint. The generation of the EOS token immediately after the $N$-th word is also considered correct.

\textbf{Failure (Too Short)}: The model generates fewer than $N$ words before the EOS token.
\textit{Example ($N=3$):} \texttt{The dog . <eos>} (2 words).
Here, generation steps for the actual words are ``correctly contributing,'' but the premature generation of the EOS token is a ``failure step'' with respect to the length constraint.

\textbf{Failure (Too Long)}: The model generates more than $N$ words before the EOS token.
\textit{Example ($N=3$):} \texttt{The dog runs fast . <eos>} (4 words).
In this case, generation steps for the first $N$ words are ``correctly contributing.'' However, any subsequent word-generating tokens (e.g., ``fast'', ``.'') are ``failure steps,'' as is the delayed EOS generation.

By analyzing DLA scores at each generation step $i$, we aim to pinpoint components whose activity strongly correlates with adherence to, or violation of, the $N$-word count instruction. We define a DLA-based metric, Cumulative Weighted Attribution (CWA), which aggregates the signed contributions of a model component $C_l$ (e.g. $A_l$, $F_l$) across the entire generation process of an output sequence $T$ of $K$ tokens (thus $K$ generation steps). Figure~\ref{fig:example} illustrates the intuition behind CWA computation in both successful and unsuccessful generations.
Formally, for a component $C_l$, its CWA is computed as
\begin{equation}
\label{eq:cwa}
  \text{CWA}_{C_l} = \frac{1}{K}\sum_{i=1}^{K}sign(t_i|t_{<i})\text{DLA}_{C_l}.
\end{equation}
where the  $sign$ function determines the sign based on whether generating token $t_i$, given what has been already generated in previous generation steps $(t_1, \dots, t_{i-1})$, correctly contributes to achieving the $N$ word count. To put it formally, let $S$ be the sentence generated from tokens $(t_1, \dots, t_{i})$ collectively constituting $m$ words. The $sign$ function can be written as
\begin{equation}
  sign(t_i|t_{<i}) = 
  \begin{cases} 
      +1 & m \leq N \\
      -1 & \text{otherwise}
   \end{cases}
\end{equation}
The normalization by $K$ allows for comparison of CWA values across different generated sequences of varying lengths. A higher positive CWA suggests the component $C_l$ consistently contributes positively to logits of tokens that lead to constraint satisfaction, while a more negative CWA suggests the opposite or contributions to failure steps. This formulation has two important properties. First, CWA is more sensitive to errors in shorter sequences. For example, generating one extra word against a target of $N=3$ results in a larger proportion of negatively-weighted steps than generating one extra word for $N=30$ . This aligns with the intuition that small absolute errors are more significant for shorter, simpler constraints. Second, CWA is a diagnostic tool, not a generation error metric. It assesses the contribution of a component at each step independently and does not model the conditional nature of generation (i.e., that the choice of token $t_i$ is highly dependent on all previous tokens $t_{<i}$). Despite this simplification, its purpose is to provide a clear, high-level signal to identify which model components are most influential in adhering to (or failing to meet) explicit length constraints.

Leveraging the CWA metric, our study proceeds with two key comparative investigations. First, we analyze differences in component contributions (via CWA scores) between foundation models and their instruction-tuned counterparts. This aims to elucidate how fine-tuning impacts component involvement when models process explicit instructions like word count constraints. Second, we extend this analysis to a multilingual setting by comparing CWA patterns across different languages. This allows us to evaluate whether the activation patterns and magnitudes of task-relevant components vary by language, thereby probing the language-specific nature of instruction-following capabilities.
\section{Experimental setup}
\label{sec:exp-setup}
This section details the experimental design employed to analyze word count control in LLMs, covering prompt construction, model selection, and the procedure for evaluating adherence to length constraints across different languages and model types.
\paragraph{Prompt construction}
We designed experiments for four settings: instruction-tuned models in English (ENG-IT) and Italian (ITA-IT), and base models in English (ENG-BASE) and Italian (ITA-BASE).
Our prompt design accounts for the fundamental differences between these model types. IT models are optimized to follow direct commands, while base BASE are trained for text completion. We therefore created two main prompt styles: (1) \textit{instructional prompts}, containing explicit commands for IT models (e.g., ``Write a sentence...''); (2) \textit{prefix prompts}, containing complete sentences for BASE models to continue (e.g., ``This is a sentence...'').
To test the robustness of each model to different phrasing, we also included a \textit{cross-condition} prompt in each set: IT models received one prefix-style prompt, and base models received one instructional prompt. All prompts were created in both English and Italian.
For the ENG-IT and ITA-IT experiments, we used the following prompt templates in conjunction with the LLM-specific chat format:
\begin{enumerate}[label= (\alph*), nosep, left=0pt]
  \small
  \item Generate a sentence using exactly \texttt{N} words.\\Genera una frase usando esattamente \texttt{N} parole.
  \item Write a text with exactly \texttt{N} words.\\Scrivi un testo con esattamente \texttt{N} parole.
  \item Write a sentence containing \texttt{N} words.\\Scrivi una frase contenente \texttt{N} parole.
  \item This is a sentence with \texttt{N} words:\\Questa è una frase con \texttt{N} parole:
\end{enumerate}
For the ENG-BASE and ITA-BASE experiments, we employed the following templates: 
\begin{enumerate}[label= (\alph*), nosep, left=0pt]
  \small
  \item This is a sentence with \texttt{N} words:\\Questa è una frase con \texttt{N} parole:
  \item This phrase has exactly \texttt{N} words from start to finish:\\Questa frase ha esattamente \texttt{N} parole dall'inizio alla fine:
  \item Here's a phrase that includes \texttt{N} words in total:\\Ecco una frase che include in totale \texttt{N} parole:
  \item Generate a sentence using exactly \texttt{N} words.\\Genera una frase usando esattamente \texttt{N} parole.
\end{enumerate} 

For each template, we substitute \texttt{N} with integers from 0 to 9, generating 10 distinct prompts per template, for a total of 80 unique prompts across all templates.
\paragraph{Models}
To ensure comparability across experiments, we adopt the Llama 3.1 8B model~\cite{grattafiori2024llama3herdmodels}, a publicly available, open-source multilingual LLM released by Meta. We use both the base version\footnote{\texttt{meta-llama/Llama-3.1-8B} on HuggingFace} and its instruction-tuned counterpart\footnote{\texttt{meta-llama/Llama-3.1-8B-Instruct} on HuggingFace}, enabling a direct comparison between foundation and instruction-tuned settings under consistent architectural conditions. 

Both versions of the model share the same architecture, consisting of 32 Transformer layers with 32 attention heads per layer. This uniformity ensures that any observed differences in behavior can be attributed to instruction tuning or language-specific effects, rather than architectural discrepancies. Furthermore, the multilingual capacity of Llama 3.1 makes it particularly well-suited for cross-linguistic comparisons in tasks requiring fine-grained control, such as word-level generation under explicit constraints.

To ensure statistical reliability, each model was tested with its appropriate prompt set, and each experimental configuration was independently repeated 20 times.
\section{Results}
\label{sec:results}
This section presents the empirical findings of our investigation. We first evaluate the baseline length control capabilities of the selected Llama 3.1 models and then delve into a detailed analysis of internal component contributions to this task using CWA.
\paragraph{Evaluating Length Control Performance}
We evaluate the length control capabilities of the selected Llama 3.1 8B model across its foundation (BASE) and instruction-tuned (IT) versions, and in both English (ENG) and Italian (ITA). This section presents these performance results, focusing on the Mean Absolute Error (MAE). The MAE quantifies the average magnitude of errors between the generated word count ($N_{\text{gen}}$) and the target word count ($N_{\text{target}}$), irrespective of their direction. For a set of $M$ generated sequences, it is calculated as:
\begin{equation}
\label{eq:mae}
\text{MAE} = \frac{1}{M} \sum_{j=1}^{M} | N_{\text{gen},j} - N_{\text{target},j} |
\end{equation}
where $N_{\text{gen},j}$ and $N_{\text{target},j}$ are the generated and target word counts for the $j$-th sequence, respectively.

\begin{figure}[t]
  \centering
  \includegraphics[width=\linewidth]{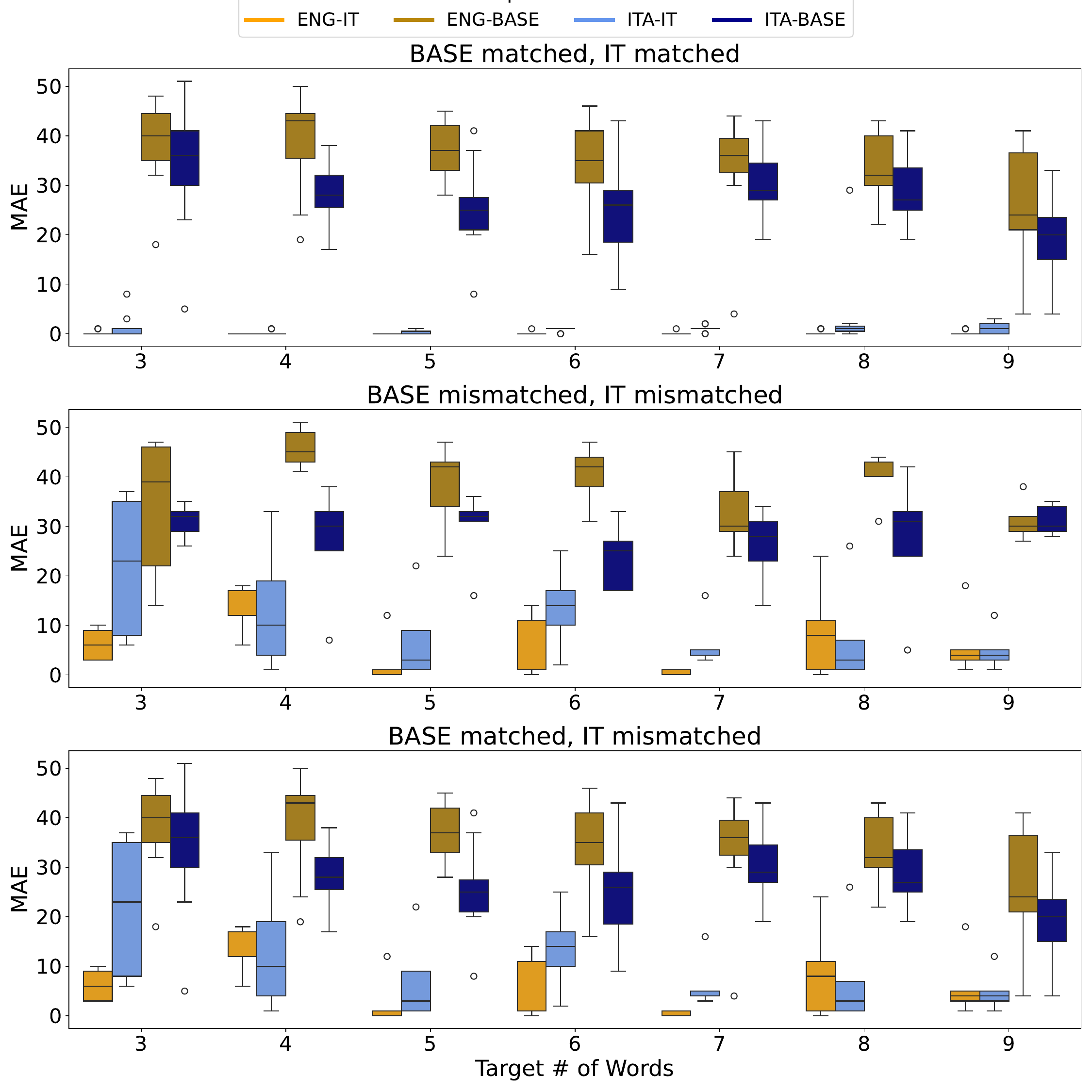}
  \caption{Error (generated word count $-$ target word count) distribution for each experiment type (ENG-IT, ENG-BASE, ITA-IT, ITA-BASE) across requested target lengths ($N \in [3,9]$). \textit{Top}: Matched prompt templates for both IT and BASE models. \textit{Middle}: Mismatched prompt templates for both IT and BASE models. \textit{Bottom}: Matched templates for BASE models and mismatched templates for IT models.}
  \label{fig:gen-error}
\end{figure}

\begin{figure}[t] 
  \centering
  \includegraphics[width=\linewidth]{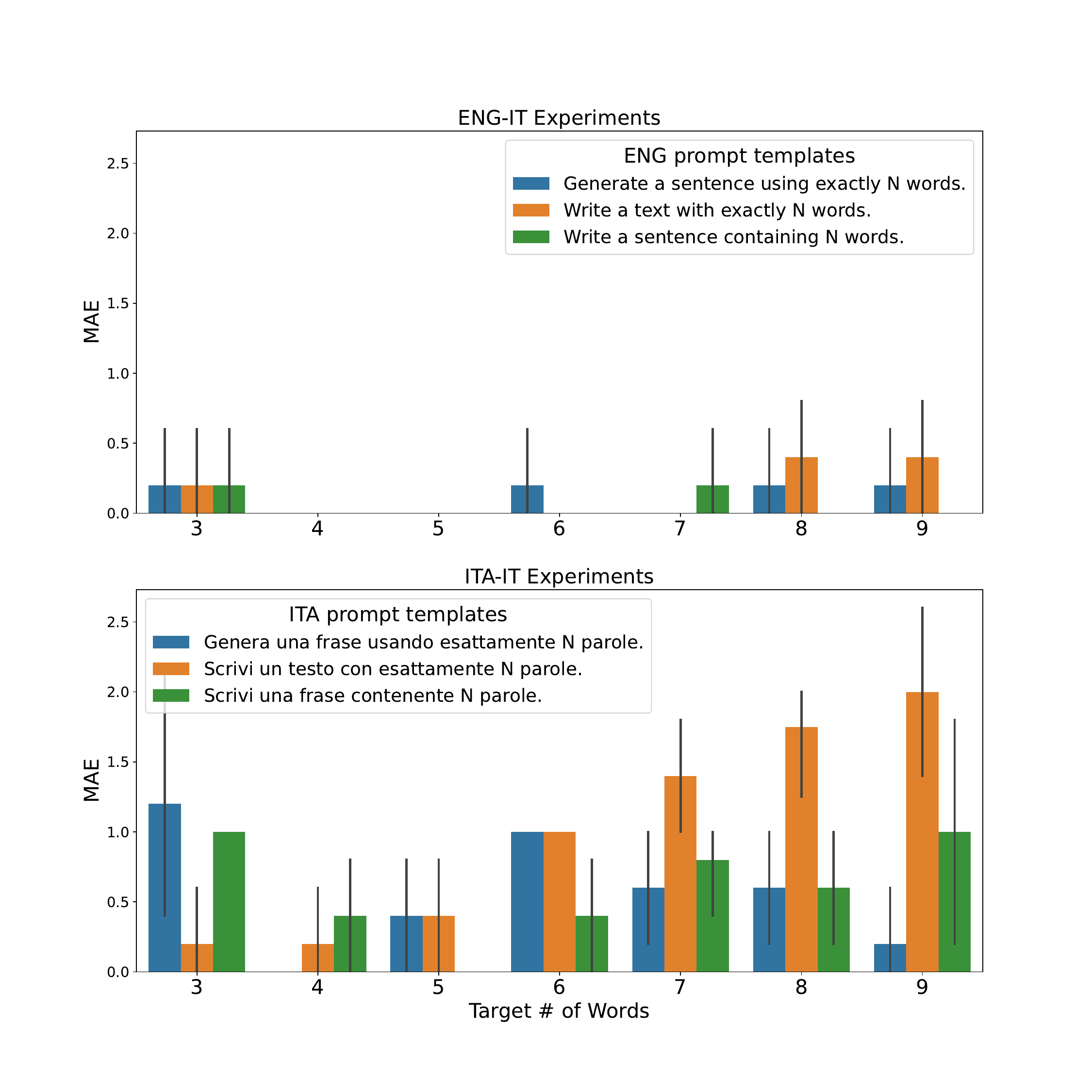}
  \caption{Mean Absolute Error (MAE) for English (ENG-IT) and Italian (ITA-IT) instruction-tuned model experiments, broken down by prompt template and target word count $N$. Outliers have been excluded in this visualization.}
\label{fig:chat-error}
\end{figure}

Figure~\ref{fig:gen-error} presents boxplots of the error distributions for each of the four experimental settings (ENG-IT, ENG-BASE, ITA-IT, ITA-BASE) across requested target lengths $N$ ranging from 3 to 9 words. The figure is divided into three subplots based on prompt template alignment:
\begin{itemize}
    \item The \textit{top subplot} shows error distributions when using prompt templates designed to match the expected input style of the LLM (i.e., instructional prompts for IT models, prefix prompts for BASE models).
    \item The \textit{middle subplot} displays error distributions when using mismatched prompting style (i.e., prefix prompts for IT models, instructional prompts for BASE models).
    \item The \textit{bottom subplot} presents a mixed scenario: matched templates for BASE models and mismatched templates for IT models.
\end{itemize}
Observing the top subplot of Figure~\ref{fig:gen-error} (matched templates), several trends emerge. Firstly, the BASE model consistently exhibits poorer length control compared to its IT counterpart. Secondly, language appears to influence performance differently for each model type: for IT models, Italian prompts seem to result in slightly worse performance than English prompts, whereas for BASE models, English prompts yield poorer results than Italian. We now focus on the performance of the IT models: Figure~\ref{fig:chat-error} illustrates the MAE for ENG-IT and ITA-IT experiments across each prompt template (a-c) and target length $N$. 
As shown, the IT model generally makes more errors with Italian templates compared to English ones. Performance also fluctuates with $N$; for example, the model appears more accurate for $N=4$ and $N=5$, aligning with some prior findings, while other target lengths present greater challenges~\cite{sun-etal-2023-evaluating}. Notably, template (b) (``Write a text with exactly N words.'', ``Scrivi un testo con esattamente N parole.'') shows a distinct pattern of decreasing performance (increasing MAE) in both languages as $N$ increases.

Table~\ref{tab:error_stats} provides descriptive statistics of the raw errors (without considering their absolute value) for each of the four experiments (ENG-IT, ENG-BASE, ITA-IT, ITA-BASE).
\begin{table}[h]
\centering
\begin{tabularx}{\linewidth}{l *{7}{>{\centering\arraybackslash}X}}
\toprule
\textbf{Experim.} & \textbf{Avg} & \textbf{Min} & \textbf{25\%} & \textbf{50\%} & \textbf{75\%} & \textbf{Max} \\
\midrule
ENG-BASE & $34.8$ $(9.45)$ & $4.0$ & $30.0$ & $36.0$ & $42.0$ & $50.0$ \\
ENG-IT   & $-0.07$ $(0.32)$ & $-1.0$ & $0.0$ & $0.0$ & $0.0$ & $1.0$ \\
\midrule
ITA-BASE & $27.3$ $(8.79)$ & $4.0$ & $21.0$ & $27.0$ & $33.0$ & $51.0$ \\
ITA-IT   & $-0.08$ $(3.12)$ & $-3.0$ & $-1.0$ & $0.0$ & $0.0$ & $29.0$ \\
\bottomrule
\end{tabularx}
\caption{Descriptive statistics of errors (generated word count $-$ target word count), grouped by language and model type. These statistics considered results where the template style matched the LLM's expected style.} 
\label{tab:error_stats}
\end{table}
Consistent with some literature~\cite{sun-etal-2023-evaluating}, the IT model tends to generate outputs that are either correct or slightly shorter than requested, as indicated by negative mean errors and third quartiles at or below 0. Overall, the IT model demonstrates strong length control for an open-source model, with a low average raw error of -0.07 for English and -0.08 for Italian experiments when using matched prompts.

\begin{figure}[h]   
  \centering
  \includegraphics[width=\linewidth]{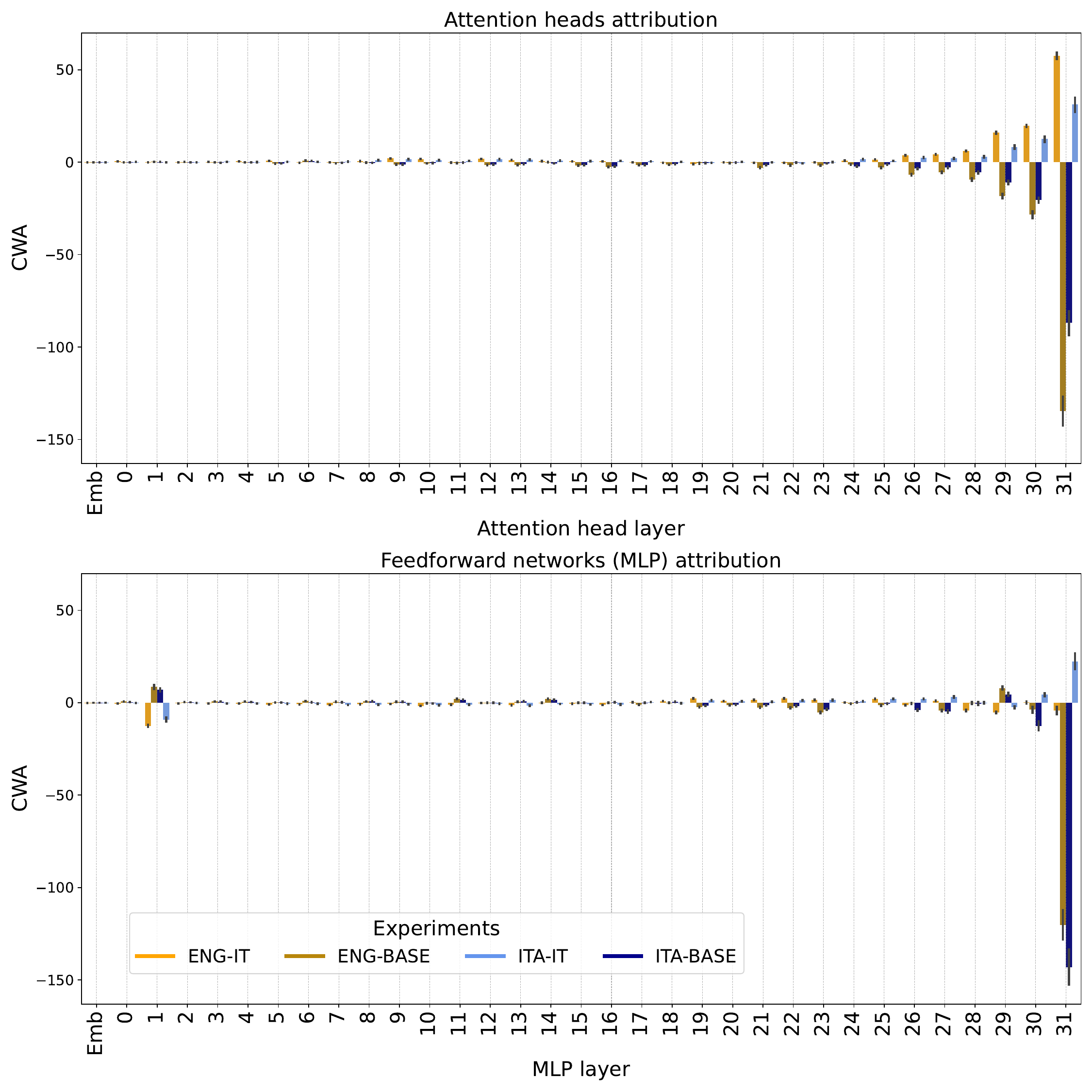}
  \caption{Cumulative Weighted Attribution (CWA) scores for (top) attention heads and (bottom) feedforward networks (MLPs) across model layers, grouped by experiment type: ENG-BASE, ENG-IT, ITA-BASE, ITA-IT. Results shown for matching prompt templates.}
  \label{fig:exp-cwa}
\end{figure}

Returning to Figure~\ref{fig:gen-error}, the middle subplot (mismatched templates) shows that even when prompted suboptimally, the IT model still makes fewer errors than the BASE model under similar mismatched conditions. The bottom subplot (BASE-matched, IT-mismatched) further reinforces this: the IT model, even with mismatched prompts, generally outperforms the BASE model using its optimal prompts. These observations suggest that the IT model actively seeks and attempts to follow instructions within the prompt, even if not formatted in its preferred style.
\paragraph{Analyzing Component Contributions to Length Control}
We begin by examining the overall CWA profiles of attention heads and MLP components across all model layers, comparing the four primary experimental settings (ENG-BASE, ENG-IT, ITA-BASE, ITA-IT) when using matched prompt templates (Figure~\ref{fig:exp-cwa}).
A striking general pattern emerges when comparing IT and BASE models, particularly in the deeper layers. Components in these later stages that exhibit positive CWA (i.e., contribute to correct length adherence) in IT model experiments often show markedly negative CWA in BASE model experiments. This suggests that if specialized mechanisms for length control are developed, they predominantly reside in the deeper layers of IT models. Conversely, a failure of these deep-layer components to contribute positively in IT models appears to negatively impact overall performance in adhering to the length constraint. An exception to this trend is observed in the MLP of layer 1, where BASE models show positive CWA while IT models exhibit negative CWA. The strong involvement of later layers in IT models for this task aligns with literature suggesting that instruction tuning primarily modifies these later layers, potentially leading to the learning of more superficial, instruction-following patterns~\cite{kung2023models, zhou2023instructionfollowingevaluationlargelanguage}.

Focusing first on the attention head contributions (Figure~\ref{fig:exp-cwa}, top), a clear positive trend in CWA for IT models begins around layer 24, with this positive attribution intensifying in subsequent, deeper layers. This effect is particularly pronounced for ENG-IT experiments, while the magnitude of positive CWA in ITA-IT experiments is somewhat attenuated. This disparity could indicate that the model, after instruction tuning, is more adept at identifying and utilizing relevant informational cues for length control from English prompts, potentially due to a stronger learned association with English instructional verbs and sentence structures, as suggested by prior work on instruction tuning~\cite{wu-etal-2024-language}. In contrast, attention heads in BASE models generally show low or negative CWA across most layers, indicating a lack of specialized attention patterns for this task.

Turning to the feedforward network (MLP) contributions (Figure~\ref{fig:exp-cwa}, bottom), the BASE model exhibits a strong and negative CWA in deeper layers. This implies that their MLP computations are largely unhelpful, or even detrimental, to satisfying the word count constraint, likely focusing instead on general language modeling. For IT models, the patterns differ by language. ENG-IT experiments show CWA values fluctuating around zero with no discernible strong trend across layers. ITA-IT experiments, however, display a notable increase in positive CWA specifically in the final layer's MLP. While the precise reasons for this distinct last-layer behavior in ITA-IT are not immediately obvious, it is observed alongside generally lower length-control performance in Italian (as noted previously). This combination of relatively lower positive attention CWA throughout the later layers and a concentrated positive MLP CWA in the very last layer for ITA-IT might contribute to the observed performance differences compared to ENG-IT.

In synthesis, these CWA patterns suggest that instruction tuning leads Llama 3.1 to specialize its later layers, particularly attention mechanisms (and for Italian, the final MLP), for responding accurately to length instructions. Deeper layers in BASE models, conversely, not only fail to demonstrate instruction-following for this specific constraint but their computations (especially MLPs) appear misaligned with the task, likely remaining focused on their pre-training objective of general language modeling.

\begin{figure}[h]
  \centering
  \includegraphics[width=\linewidth]{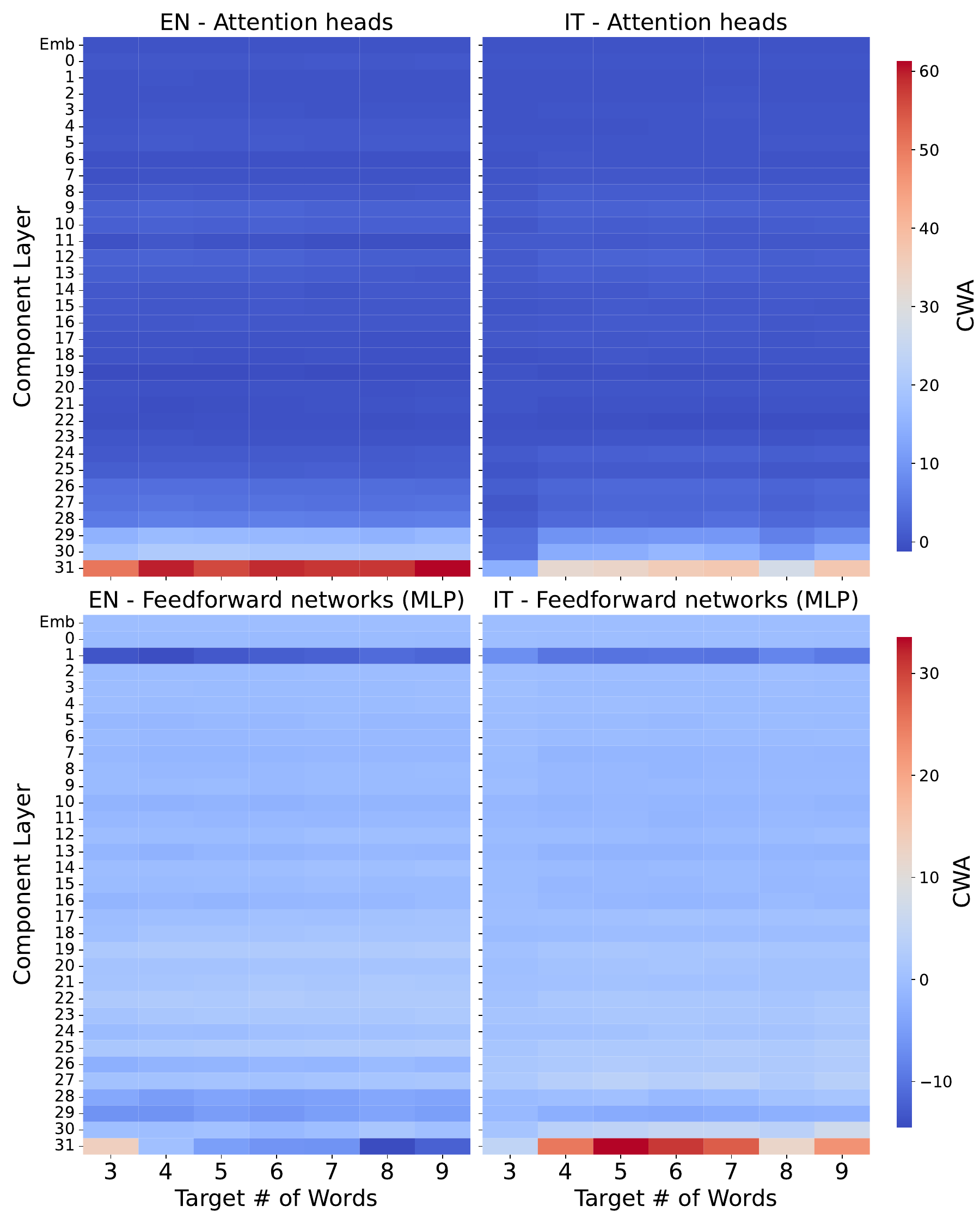}
  \caption{Heatmaps of CWA scores for instruction-tuned models (ENG-IT left, ITA-IT right), using matched prompts. \textit{Top row}: Attention head CWA. \textit{Bottom row}: MLP component CWA. Scores are shown across model layers (y-axis) and varying target word counts $N$ (x-axis, from 3 to 9 words).}
  \label{fig:ntokens-cwa}
\end{figure}

Further insights into component involvement are revealed by examining CWA scores as a function of the requested number of words ($N \in [3,9]$). This analysis focuses on the IT experiments, comparing ENG-IT and ITA-IT when using matched prompt templates. Figure~\ref{fig:ntokens-cwa} presents these results as heatmaps, illustrating CWA values for attention heads (top row) and MLP components (bottom row) across model layers (y-axis) and varying target word counts $N$ (x-axis). The left column corresponds to ENG-IT experiments, and the right column to ITA-IT experiments.
Considering the attention head contributions (Figure~\ref{fig:ntokens-cwa}, top row), the heatmaps corroborate earlier findings: ITA-IT experiments generally exhibit lower positive CWA values across most layers and target lengths compared to ENG-IT. A particularly noticeable dip in CWA for ITA-IT occurs when $N=3$. Interestingly, $N=3$ also corresponds to the lowest average attention CWA in the later layers for ENG-IT experiments. This reduced positive attention CWA at $N=3$ for both languages, especially in deeper layers crucial for instruction following, might suggest that attention mechanisms struggle more to extract or route length-specific information when the $N=3$, potentially contributing to the poorer length control performance observed (as discussed previously).

Examining the MLP contributions (Figure~\ref{fig:ntokens-cwa}, bottom row), a striking contrast emerges between ENG-IT and ITA-IT, particularly in the final layers. For ITA-IT, later-layer MLPs often show stronger positive CWA values across various target word counts $N$. Conversely, for ENG-IT, the corresponding MLPs in the same final layers frequently exhibit negative CWA or values close to zero. This observation, particularly the positive CWA in late-layer MLPs for ITA-IT, might indicate a compensatory mechanism: if attention heads in Italian are less effective at isolating and propagating length-relevant information (as suggested by their generally lower CWA), the MLPs in subsequent layers, especially the final ones, might undertake more intensive processing to meet the constraint. This increased reliance on later-layer MLP processing for Italian could be a factor in the observed differences in overall length control performance between the two languages.
\section{Conclusions}
\label{sec:conclusion}

This study investigated the impact of instruction-tuning on a Large Language Model's ability to adhere to precise word count constraints. By comparing the foundation (BASE) and instruction-tuned (IT) variants of the Llama 3.1 8B model in English and Italian, we confirmed that instruction-tuning provides a substantial performance boost. The IT model demonstrated strong length control, whereas the BASE model consistently failed, typically by over-generating text. However, we noted a slight degradation in the IT model's accuracy when prompted in Italian compared to English, indicating that instruction-following capabilities can be language-dependent.

To understand the mechanisms driving these differences, we introduced Cumulative Weighted Attribution (CWA), a metric derived from direct logit attribution. Our interpretability analysis revealed that instruction-tuning fundamentally reshapes the model's deeper layers to handle the task. Components in these later layers of the IT model consistently showed positive CWA scores, signaling their beneficial contribution to satisfying the length constraint. In contrast, the same components in the BASE model often exhibited negative CWA scores, indicating their activity was detrimental to the task.

Delving deeper, we identified later-layer attention heads as particularly influential in the IT model, especially in English. We hypothesize that these components become specialized at tracking and implementing the length constraint. In Italian, where attention contributions were more attenuated, we observed a stronger compensatory role from final-layer MLPs. This suggests that the model develops flexible, language-specific strategies for task adherence, reallocating computational responsibility between components based on context.

\paragraph{Limitations}
Our study has several limitations. First, our experiments used only single-digit numbers for the target word count, which may not expose challenges related to parsing multi-token numbers. Second, our analysis was confined to the Llama 3.1 8B architecture; these findings may not generalize to larger models or different model families (e.g., Mistral, Gemma), which could exhibit different component specializations. Finally, while we explored English and Italian, a broader set of languages is needed to form a comprehensive cross-lingual understanding of these mechanisms.

\paragraph{Future Work}
These limitations pave the way for future investigations. A promising direction is to employ causal interventions, such as activating or suppressing the identified components, to move beyond correlational analysis and definitively establish their functional roles in length control. Furthermore, our analytical framework could be extended to other forms of controlled generation, such as style transfer or sentiment control, to determine if similar component-level specializations emerge across different explicit constraints.
\section*{Acknowledgments}
This work was supported in part by project SERICS (PE00000014) under the NRRP MUR program funded by the EU - NGEU. Views and opinions expressed are however those of the authors only and do not necessarily reflect those of the European Union or the Italian MUR. Neither the European Union nor the Italian MUR can be held responsible for them.


\bibliography{main}




\end{document}